\pdfoutput=1
\PassOptionsToPackage{table}{xcolor}
\documentclass[sigconf,natbib=true,nonacm]{acmart}
\usepackage{etoolbox}
\makeatletter
\patchcmd{\@mkauthors@iii}{\par\medskip\leavevmode}{\par\vskip 14pt\leavevmode}{}{\PackageWarning{main}{author top space patch failed}}
\patchcmd{\@mkauthors@iii}{\and\par\bigskip}{\and\par\vskip 20pt}{}{\PackageWarning{main}{author bottom space patch failed}}
\makeatother

\AtBeginDocument{%
  \providecommand\BibTeX{{%
    \normalfont B\kern-0.5em{\scshape i\kern-0.25em b}\kern-0.8em\TeX}}}

\usepackage{xspace}
\usepackage{comment}
\usepackage[noabbrev,capitalise]{cleveref}
\usepackage{algorithm}
\usepackage{algorithmic}
\usepackage{multirow}
\usepackage{balance}
\usepackage{subfigure}
\usepackage{graphicx}
\usepackage{mdframed}
\usepackage{tcolorbox}
\usepackage{xcolor}
\usepackage{mathtools}
\usepackage{enumitem}

\usepackage{amsmath}
\usepackage{amsthm}

\usepackage{ragged2e}
\usepackage{colortbl}
\usepackage[table]{xcolor}

\definecolor{lightblue}{RGB}{173,216,230}
\definecolor{lightgreen}{RGB}{144,238,144}

\newcommand{\nosection}[1]{\vspace{2pt}\noindent\textbf{#1.}}

\newcommand{\ourmodel}{\textbf{JevRec}\xspace}
\crefname{equation}{Eq.}{Eqs.}
\Crefname{equation}{Eq.}{Eqs.}
\crefname{appendix}{Appendix}{appendices}
\Crefname{appendix}{Appendix}{Appendices}

\begin{document}

\title{JevRec: Linguistic Encoding with Jev-like Decision Modeling for LLM-based Generative Recommendation}

\author{Qiyong Zhong}
\affiliation[obeypunctuation=true]{%
  \institution{Zhejiang University},
  \city{Hangzhou},
  \country{China}}
\email{youngzhong@zju.edu.cn}

\author{Jiajie Su}
\affiliation[obeypunctuation=true]{%
  \institution{Zhejiang University},
  \city{Hangzhou},
  \country{China}}
\email{sujiajie@zju.edu.cn}

\author{Ming Yang}
\affiliation[obeypunctuation=true]{%
  \institution{Zhejiang University},
  \city{Hangzhou},
  \country{China}}
\email{yummyyummy183@gmail.com}

\author{Li Zhang}
\affiliation[obeypunctuation=true]{%
  \institution{Zhejiang University},
  \city{Hangzhou},
  \country{China}}
\email{zhanglizl80@gmail.com}



\begin{abstract}
Generative recommendation (GR) represents items as semantic identifiers (SIDs) and autoregressively generates the next item's SID from interaction history. However, most methods assign the same SID length to all items despite highly unequal interaction frequencies. This uniform design repeatedly incurs the cost of fine-grained codes for popular items, while shortening codes to reduce this cost also compresses the semantic representations of long-tail items. Inspired by the compact forms of frequent expressions in natural language, we propose \ourmodel to encode items as variable-length words and generate the next item from interaction context. Linguistic encoding retains frequency-dependent semantic prefixes from shared codebooks, producing shorter SIDs for popular items and richer descriptions for long-tail items. Yet predicting a sequence-level length and generating token-level content impose different optimization requirements that may conflict under a shared output head. Our Jev-like decision modeling component therefore predicts a length distribution with a dedicated head and generates content conditioned on length through a separate head while sharing the interaction-history representation. Experiments on three public datasets show that \ourmodel outperforms all compared baselines across the evaluated recommendation metrics. Further analyses show lower training and inference time than TIGER and reduced gradient conflict between length and content prediction.
\end{abstract}

\maketitle

\setlength{\floatsep}{4pt plus 4pt minus 1pt}
\setlength{\textfloatsep}{4pt plus 2pt minus 2pt}
\setlength{\intextsep}{4pt plus 2pt minus 2pt}
\setlength{\dbltextfloatsep}{3pt plus 2pt minus 1pt}
\setlength{\dblfloatsep}{3pt plus 2pt minus 1pt}
\setlength{\abovecaptionskip}{3pt}
\setlength{\belowcaptionskip}{2pt}
\setlength{\abovedisplayskip}{2pt plus 1pt minus 1pt}
\setlength{\belowdisplayskip}{2pt plus 1pt minus 1pt}

\section{Introduction}
\label{sec:introduction}

\begin{figure}[t]
    \centering
    \includegraphics[width=\columnwidth]{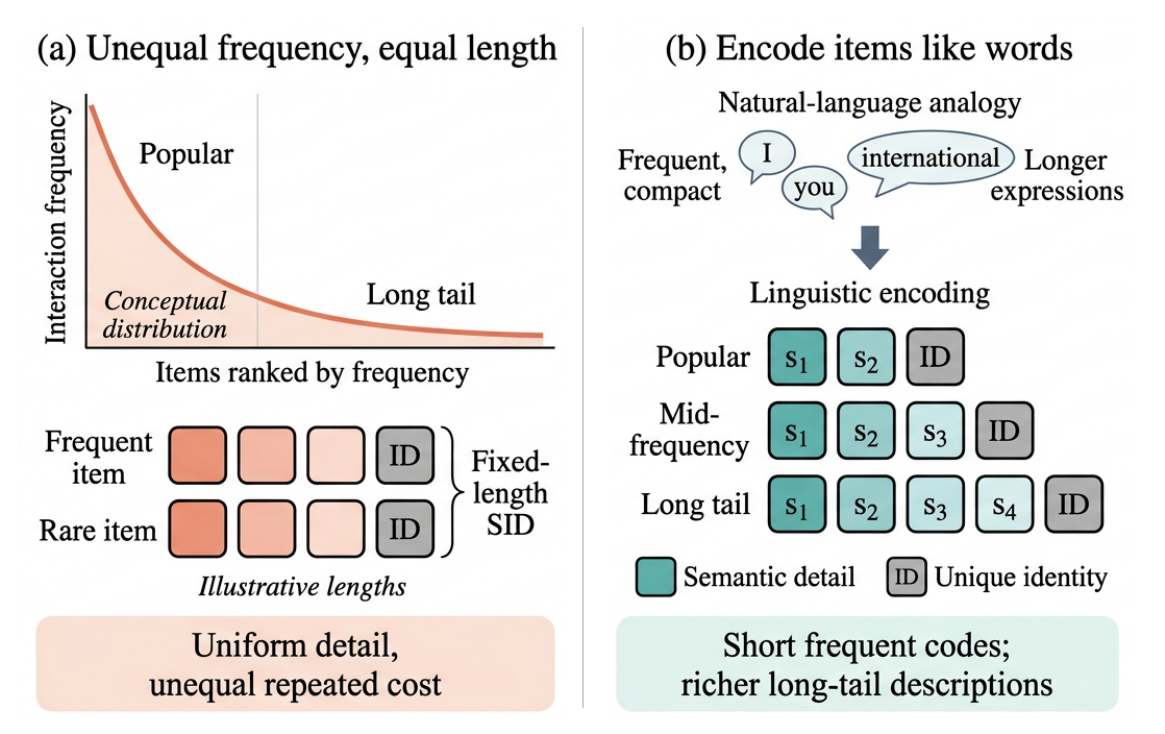}
    \caption{\textbf{Motivation for frequency-aware SID lengths.} (a) Fixed-length codes impose equal representation lengths despite unequal interaction frequencies. (b) The natural-language analogy motivates shorter codes for popular items and richer descriptions for long-tail items. The distribution and code examples are illustrative.}
    \label{fig:motivation}
    \Description{Two panels illustrate the motivation. The left panel shows a conceptual long-tailed interaction-frequency curve and equal-length codes for frequent and rare items. The right panel compares short frequent words with longer expressions, then shows item codes whose semantic part grows from popular to long-tail items while retaining an identity token.}
\end{figure}

Generative recommendation (GR)~\citep{tiger,deng2025onerec} represents each item as a sequence of discrete semantic ID (SID) tokens and autoregressively generates the next item's SID from the user's interaction history, following the next-token generation paradigm of large language models (LLMs)~\citep{sod,speeder}. Unlike conventional recommendation methods~\citep{gru4rec,sasrec} that maintain a separate ID embedding for every item, GR composes item representations from a compact, shared token vocabulary, reducing the storage burden of item embedding tables as the catalog grows and offering advantages in memory efficiency~\citep{hou2025rpg} and scalability~\citep{hstu,actionpiece}. Residual quantization~\citep{rqvae} maps item content to SIDs whose shared prefixes allow semantically similar items to share knowledge~\citep{tiger}, enabling recommendation directly in the code space. GR has attracted increasing research attention, with recent efforts improving item tokenization~\citep{letter,liu2025e2egrec} or assigning different SIDs to the same item in different user contexts~\citep{zhong2025pctx}. However, most methods assign SIDs of the same length to all items~\citep{tiger,letter}, such as three semantic codes followed by a collision-resolution code.

Fixed-length SIDs tie the semantic representation requirements of the entire item catalog to the encoding cost incurred under the actual interaction distribution through a single length. Recommendation data typically follow a long-tailed distribution in which many low-frequency items expand the catalog while a small set of popular items accounts for a large share of interactions. Consequently, adding one semantic code incurs different cumulative costs for items with different interaction frequencies. A uniform code length cannot exploit this difference: the additional cost of retaining fine-grained semantics is repeatedly paid on frequent interactions, whereas reducing that cost requires compressing all items together. As illustrated in~\Cref{fig:motivation}, variable-length SIDs offer a way to move beyond this uniform constraint and balance semantic detail against encoding cost according to item frequency.

Natural language accommodates words of different lengths and often favors simpler forms for frequent expressions, such as shorter words or fewer written strokes. For example, the English words \emph{I} and \emph{you} require only a few letters, whereas \emph{international} and \emph{communication} have longer forms. The useful principle is not to shorten every word, but to reduce the cost of frequent expressions while preserving a rich vocabulary. Length variation also provides an additional cue for distinguishing word forms, suggesting that a uniform-length vocabulary could be harder to learn without this cue. In this work, we explore modeling item SIDs as word-like units, viewing the generation of the next SID from interaction history as analogous to how humans formulate language in context. We introduce \textbf{linguistic encoding} to give item representations frequency-dependent lengths: shorter SIDs for popular items and longer semantic descriptions for long-tail items. This analogy motivates both lower training cost and faster convergence in GR by shortening frequent representations and providing length-based cues for learning the item vocabulary.

Specifically, we group items by interaction frequency and retain different code lengths within a shared set of semantic codebooks: popular items discard trailing fine-grained codes, while long-tail items retain their full semantic descriptions. We append collision-resolution codes within each group to preserve unique item identities after truncation. Under this encoding, length is a global property of the entire SID rather than an ordinary content symbol. Length decision-making focuses on sequence-level classification, whereas content generation models fine-grained semantic dependencies conditioned on the generated prefix, so their requirements for discriminative features and parameter updates need not align. A shared generation head couples these capabilities through a common output mapping and may introduce optimization competition: updates that improve local token prediction do not necessarily improve global length classification. Moreover, each SID provides only one length label but multiple content-token labels, which may bias joint optimization toward content generation. We therefore introduce \textbf{Jev-like decision modeling}~\citep{jev1,jev2,jev3} with task-specific prediction heads for length classification and content generation to alleviate their coupling and competition at the output layer. The length head estimates the probabilities of candidate lengths, while the content head generates item codes conditioned on length, separating their roles while sharing the interaction-history representation.

We conduct extensive experiments on three public datasets, with results showing that \ourmodel outperforms all compared baselines on every evaluated recommendation metric. Our efficiency analysis also shows shorter training time and faster convergence, supporting the computational and learning benefits of the language-inspired design.

Our main contributions are summarized as follows:
\begin{itemize}[leftmargin=*,topsep=0pt]
    \item \textbf{Linguistic encoding.} Inspired by compact frequent expressions in natural language, we give popular items shorter codes and long-tail items longer semantic descriptions, allowing SID length and semantic granularity to vary with item frequency. The codes preserve shared semantics and unique item identities while reducing repeated encoding costs for frequent interactions.
    \item \textbf{Jev-like decision modeling.} We introduce task-specific heads for length classification and content generation to alleviate competition in a shared output layer. The heads share history representations and interact through length-conditioned generation, with length probabilities guiding candidate selection across decoding branches to form a length-aware recommendation list.
\end{itemize}

\section{Related Work}
\label{sec:related}

\nosection{Generative Recommendation} Sequential recommendation~\citep{lrprec} has progressed from models of item transitions to attention-based history encoders~\citep{rendle2010fpmc,sasrec,bert4rec}, with further developments incorporating multiple behavior types and multimodal signals~\citep{su1,su2,su3}. SID-based generative recommendation follows a different retrieval formulation: a model predicts a discrete identifier sequence instead of scoring a separate embedding for every item~\citep{tay2022dsi,tiger}. A shared token vocabulary supports compact item representations and knowledge reuse~\citep{tiger,mbrec}, while semantic identifiers also connect recommendation with large language models (LLMs)~\citep{masopd,usa}. Large-scale sequence models and industrial deployments further demonstrate the practical relevance of generative approaches~\citep{hstu,yan2025lum,deng2025onerec,mtgr2025,he2025plum}. Item tokenization is a central design choice in this paradigm. Existing constructions include learned quantization~\citep{tiger,letter,zhu2024cost,hou2025rpg}, cluster-derived indices~\citep{tay2022dsi,hua2023index,wang2024eager}, identifiers produced with language models~\citep{jin2023language,liu2025e2egrec}, and quantizers that integrate multiple modalities~\citep{xu2025mmq}. Joint optimization of the tokenizer and recommender seeks to adapt the code space to recommendation objectives~\citep{liu2025e2egrec,li2026unigrec}. Other studies examine how hierarchical identifiers constrain prediction probabilities~\citep{hou2026latte}, how compression affects semantic information~\citep{ma2026acerec,hou2025rpg}, and whether token prediction encourages memorization rather than generalization~\citep{ding2026generalize}. Beyond code construction, recent work investigates diffusion-based generation~\citep{shi2025lladarec,liu2025diffgrm}, multi-level alignment~\citep{ye2025align3gr}, multimodal recommendation~\citep{liu2024mmgrec,zhu2025mgr}, next-place prediction~\citep{wang2025generative}, and retrieval-assisted generation for unseen items~\citep{ding2026specgr}. ActionPiece~\citep{actionpiece} also incorporates action context into tokenization. Pctx~\citep{zhong2025pctx} personalizes item tokenization using user interaction history. Our work aims to allocate SID lengths according to interaction frequency and coordinate sequence-level length decisions with token-level content generation.

\nosection{Tokenization} A tokenizer defines the discrete units through which a model observes and generates data. Its granularity determines both the number of symbols needed to represent an input and the size of the vocabulary from which those symbols are selected. In language processing, BPE and WordPiece construct reusable subword vocabularies~\citep{sennrich2016bpe,wu2016wordpiece}, whereas unigram tokenization uses a probabilistic segmentation model~\citep{kudo2018unigram}; SentencePiece provides a language-independent framework for subword processing~\citep{sentencepiece}. Visual tokenizers instead learn discrete latent codes from encoded image representations~\citep{van2017neural,esser2021taming,yu2024image}. These approaches illustrate how token design mediates between representational detail and sequence length. Semantic item tokenizers transfer discrete representation learning to recommendation~\citep{tiger,deng2025onerec}, but the frequency of an item determines how often its encoding cost is incurred. Recent studies also explore variable-length item representations~\citep{SA2CRQ,VSID,BONSAI,SST}. These approaches emphasize adaptive encoding, history compression, or decoding-tree structure, whereas we coordinate frequency-aware encoding with explicit target-length decisions during generation. Motivated by compact frequent expressions in language, \ourmodel varies the length of complete item codes rather than imposing a uniform semantic budget. Linguistic encoding retains shorter prefixes for popular items and finer-grained descriptions for long-tail items, while Jev-like decision modeling makes length an explicit prediction task that conditions subsequent content generation.

\section{Method}
\label{sec:method}
\subsection{Overview}
\label{sec:method-overview}

\begin{figure*}[t]
    \centering
    \includegraphics[width=0.88\textwidth]{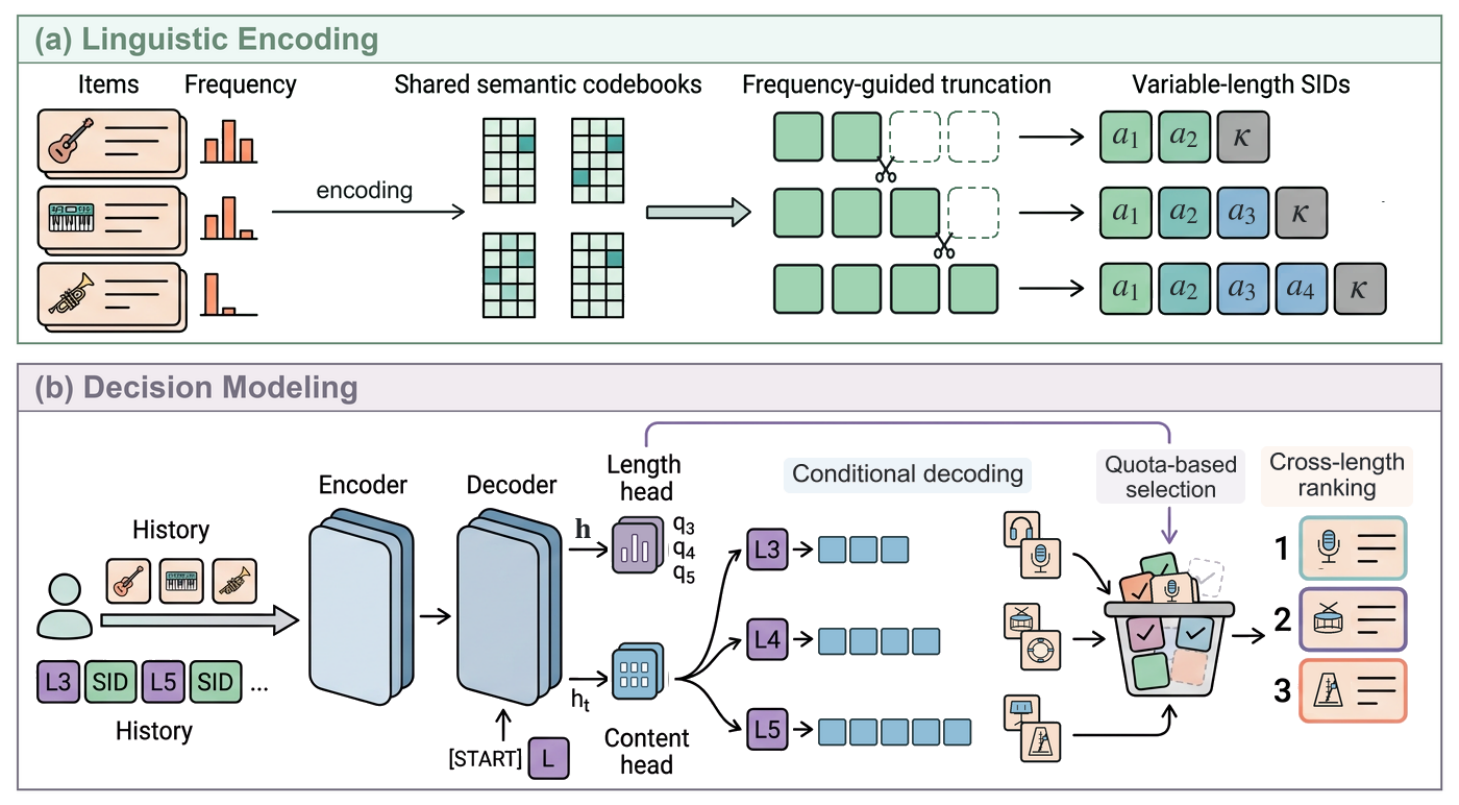}
    \caption{\textbf{Overview of \ourmodel.} (a) Linguistic encoding constructs frequency-dependent SIDs with shared codebooks. (b) Jev-like decision modeling separates length and content prediction, followed by conditional decoding, quota-based selection, and cross-length ranking. The blue blocks in (b) represent complete SIDs; item examples are illustrative.}
    \label{fig:framework}
    \Description{The upper panel encodes item text with shared semantic codebooks, truncates codes by item frequency, and appends a collision token. The lower panel encodes an interaction history and uses separate length and content heads. Length-conditioned branches generate complete SIDs, which map to candidate items before quota-based selection and cross-length ranking.}
\end{figure*}

As illustrated in~\Cref{fig:framework}, \ourmodel combines frequency-aware item encoding with explicit length decision-making. Linguistic encoding assigns each item a unique SID whose length reflects its interaction frequency: popular items retain short semantic prefixes, whereas long-tail items retain finer-grained descriptions. The recommendation model then separates length classification from content generation while sharing the representation of user history. A candidate-conditioned decision head estimates a distribution over SID lengths, and a content head generates item codes under a specified length prefix. We train both heads jointly and use the length distribution to allocate candidate slots across decoding branches before ranking the selected items. The following sections define the task and detail the encoding, decision, training, and inference procedures.

\subsection{Problem Formulation}
\label{sec:problem}

Let $\mathcal V$ denote the item catalog and $S=[v_1,\ldots,v_n]$ a user's chronological interaction history, where $v_i\in\mathcal V$. Sequential recommendation predicts the next item $v^\star\in\mathcal V$ given $S$. Each item has a text description and frequency $f(v)$ computed from training interactions only. We represent an item by a discrete SID $\mathbf c(v)=(c_1(v),\ldots,c_{\ell_v}(v))$, where $\ell_v$ may vary across items. The context-invariant mapping $v\mapsto\mathbf c(v)$ uniquely identifies each item.

For a configured length set $\mathcal A$, let $L_\ell$ be the special token denoting length $\ell\in\mathcal A$. We serialize the history as
\begin{equation}
    X=\mathop{\mathrm{concat}}_{i=1}^{n}
    [L_{\ell_{v_i}},\mathbf c(v_i)].
    \label{eq:history-serialization}
\end{equation}
No additional separator is inserted between consecutive items. A length prefix specifies how many subsequent tokens belong to that item's SID. Throughout the paper, SID length counts the semantic codes and the collision-resolution code, but excludes the length token, decoder start token, and end-of-sequence token. Given $X$, the model predicts the target length $\ell_{v^\star}$ and generates $\mathbf c(v^\star)$ followed by $\mathtt{EOS}$. Recommendation returns an ordered list of $K$ distinct items recovered from generated SIDs.

\subsection{Linguistic Encoding}
\label{sec:linguistic-encoding}

\nosection{Frequency-dependent lengths}
We order items by their training frequencies $f(v)$ and divide the catalog into $N$ frequency groups, with group $1$ containing the highest-frequency items. Let $g(v)\in\{1,\ldots,N\}$ be the group of item $v$ and $M\geq2$ the minimum SID length. We assign
\begin{equation}
\begin{aligned}
    \ell_v&=M+g(v)-1,\\
    \mathcal A&=\{M,M+1,\ldots,M+N-1\}.
\end{aligned}
\label{eq:frequency-length}
\end{equation}
This assigns fewer tokens to frequently occurring items while allowing less frequent items to retain more semantic detail. The group sizes follow the relative popularity of items rather than their absolute interaction counts, so that every group remains sufficiently populated under long-tailed distributions. The partition is computed once from training frequencies and does not depend on the current user.

\nosection{Shared semantic quantization}
We encode each item's textual description with a pretrained text encoder and apply principal component analysis with whitening to obtain $\mathbf x_v\in\mathbb R^{d_e}$. A single residual quantizer is fitted to these item representations using $Q=M+N-2$ codebooks, each containing $B$ code vectors. Let $\boldsymbol\mu_{j,b}$ be entry $b$ of codebook $j$. Starting from $\mathbf r_0(v)=\mathbf x_v$, residual quantization assigns
\begin{equation}
\begin{aligned}
    a_j(v)&=\mathop{\arg\min}_{b\in\{1,\ldots,B\}}
       \|\mathbf r_{j-1}(v)-\boldsymbol\mu_{j,b}\|_2^2,\\
    \mathbf r_j(v)&=\mathbf r_{j-1}(v)-\boldsymbol\mu_{j,a_j(v)},
    \quad j=1,\ldots,Q.
\end{aligned}
\label{eq:residual-quantization}
\end{equation}
All frequency groups share these codebooks. The resulting ordered codes describe an item's content through successive residual refinements, so truncation changes the amount of retained detail without learning a separate semantic space for each group.

\nosection{Truncation and unique identification}
An item in group $g$ retains the first $d_g=M+g-2=\ell_v-1$ semantic codes and discards the remaining $N-g$ codes. We then append a collision-resolution token $\kappa(v)$ to obtain
\begin{equation}
    \mathbf c(v)=\big(a_1(v),\ldots,a_{d_{g(v)}}(v),\kappa(v)\big).
    \label{eq:variable-sid}
\end{equation}
Here each semantic index is represented by a token that preserves its codebook identity, and collision tokens belong to a vocabulary distinct from the semantic tokens. Within each group, items sharing the same retained semantic prefix receive different collision tokens. Collision indices can be reused across different prefixes because identification depends on the full SID. We allocate sufficient collision-token capacity to distinguish every item in each prefix bucket, which yields one unique SID per item without requiring cross-group collision handling.

With the default $M=N=3$, the shared quantizer has four levels. The high-, medium-, and low-frequency groups retain two, three, and four semantic codes, respectively, producing SID lengths of three, four, and five after adding the collision token. Thus lengths vary through semantic truncation without losing item identity.

\subsection{Jev-like Decision Modeling}
\label{sec:decision-modeling}

Length prediction is a sequence-level decision over a small candidate set, whereas content generation predicts successive tokens from an evolving prefix. Coupling them in one vocabulary output head can create competition between these different prediction requirements. We instead give length classification a dedicated readout and condition content generation on its chosen length, while preserving a shared representation of user history.

\nosection{History and length representations}
A shared encoder--decoder processes the serialized history $X$. We use the decoder's final-layer state at its initial position as the history-conditioned decision state:
\begin{equation}
\begin{aligned}
    H&=\operatorname{Encoder}_{\theta}(X),\\
    \mathbf h&=\operatorname{Decoder}_{\theta}
        ([\mathtt{START}],H)_0.
\end{aligned}
\label{eq:decision-state}
\end{equation}
The state $\mathbf h$ has already attended to the encoded history and is not the raw start-token embedding. For each $\ell\in\mathcal A$, we obtain a candidate representation $\mathbf e_\ell=E[L_\ell]$ from the model's token embedding table $E$. The same embedding is used when $L_\ell$ serves as a prefix for content generation, tying the decision candidate to its downstream conditioning signal.

\nosection{Candidate-conditioned scoring}
We score every legal length with one shared nonlinear function:
\begin{equation}
\begin{aligned}
    z_\ell&=f_\phi([\mathbf h;\mathbf e_\ell]),\\
    q(\ell\mid X)&=\frac{\exp(z_\ell)}
       {\sum_{t\in\mathcal A}\exp(z_t)}.
\end{aligned}
\label{eq:length-decision}
\end{equation}
The scorer $f_\phi$ is a shared three-layer MLP applied to the concatenated history and length representations, with architecture details in~\Cref{app:implementation}. Nonlinear interaction lets each candidate's score depend on the user history rather than merely adding a history-independent length bias. The head normalizes only over legal lengths instead of the full content vocabulary. Reusing $\mathbf h$ across scores requires one history encoding and one decoder start position rather than a separate model pass per candidate.

\nosection{Length-conditioned content generation}
For a candidate length $\ell$, the content head starts from $[\mathtt{START},L_\ell]$ and autoregressively generates the SID followed by $\mathtt{EOS}$. Let $\mathbf y(v)=[\mathbf c(v),\mathtt{EOS}]$. For an item whose assigned length is $\ell$, its conditional code probability is
\begin{equation}
    r_\ell(v\mid X)
    =\prod_{j=1}^{\ell+1}
      p\big(y_j(v)\mid X,\ell,\mathbf y_{<j}(v)\big).
    \label{eq:conditional-code-probability}
\end{equation}
The length and content heads together assign the joint probability
\begin{equation}
    p(\ell_v,\mathbf y(v)\mid X)
    =q(\ell_v\mid X)\,r_{\ell_v}(v\mid X).
    \label{eq:joint-code-probability}
\end{equation}
The length decision supplies the branch probability, while the content head distinguishes items under that length. This separation provides task-specific output mechanisms without removing shared learning through the history representation or the length embeddings. At inference we retain multiple length branches instead of only the most probable one, as detailed in~\Cref{sec:length-aware-inference}.

\subsection{Joint Training}
\label{sec:joint-training}

For a training pair $(S,v^\star)$, let $\ell^\star=\ell_{v^\star}$ and $\mathbf y^\star=[\mathbf c(v^\star),\mathtt{EOS}]$. We supervise the length head with $\ell^\star$ and train the content head under the corresponding ground-truth prefix $L_{\ell^\star}$. The true length aligns the conditioning prefix with the target code, giving the content head a well-defined conditional generation task. Each example contributes a content loss only in its true-length branch.

With teacher forcing, a single decoder pass receives the start token followed by the true length prefix and all tokens of $\mathbf c(v^\star)$ in order. The start-position state supplies the length decision, the length-prefix state predicts the first SID token, and the final SID position predicts $\mathtt{EOS}$. Causal attention prevents the start position from accessing the true length prefix or subsequent target tokens. The content head is not trained to predict the length token itself.

For a minibatch $\mathcal B$, the joint objective is
\begin{equation}
\mathcal L=-\frac{1}{|\mathcal B|}
\sum_{(X,\mathbf y^\star)\in\mathcal B}
\log p(\ell^\star,\mathbf y^\star\mid X).
\label{eq:joint-loss}
\end{equation}
We jointly optimize the length and content losses with unit weights in one stage. Content-token losses exclude padding and are summed per example before averaging over the batch. This objective is the negative log-likelihood of the target length and its EOS-terminated code under the decomposition in~\Cref{eq:joint-code-probability}. Both objectives update the shared backbone to learn history representations for length selection and fine-grained item discrimination.

\subsection{Length-Aware Inference}
\label{sec:length-aware-inference}

\nosection{Candidate allocation}
Rather than committing to one predicted length, we retain the full decision distribution $q(\ell\mid X)$ and generate candidates for each legal length. For a recommendation budget $K$, the largest-remainder rule converts length probabilities into integer quotas. We first compute
\begin{equation}
\begin{aligned}
    b_\ell&=\lfloor Kq(\ell\mid X)\rfloor,\\
    \rho_\ell&=Kq(\ell\mid X)-b_\ell,\\
    R&=K-\sum_{\ell\in\mathcal A}b_\ell.
\end{aligned}
\label{eq:length-quotas}
\end{equation}
Starting from $n_\ell=b_\ell$, we give one additional slot to each of the $R$ lengths with the largest $\rho_\ell$, so that $\sum_\ell n_\ell=K$. Ties favor larger length probabilities, with any remaining ties broken using a fixed random seed. Quotas determine the composition of the candidate list rather than the final ordering of its items.

\nosection{Conditional decoding}
Each branch performs beam search with width $K$ from $[\mathtt{START},L_\ell]$ while reusing the encoded history. We retain completed codes only when they map to catalog items of the corresponding SID length and select up to $n_\ell$ distinct items with the highest conditional probabilities $r_\ell(v\mid X)$ from~\Cref{eq:conditional-code-probability}. Invalid codes do not consume recommendation slots. Decoding does not use a catalog-prefix trie: candidate probabilities use the model's original token probabilities, including $\mathtt{EOS}$, without length normalization or renormalization over valid catalog prefixes.

\nosection{Cross-length ranking}
The selected candidates are merged and sorted by the joint score
\begin{equation}
\begin{aligned}
    s(v\mid X)&=q(\ell_v\mid X)\,r_{\ell_v}(v\mid X),\\
    \log s(v\mid X)&=\log q(\ell_v\mid X)
        +\log r_{\ell_v}(v\mid X).
\end{aligned}
\label{eq:inference-score}
\end{equation}
The length factor supplies the probability of entering a branch, while the conditional factor distinguishes items within that branch. This score uses the same decomposition as training and permits candidates from different lengths to interleave in the final ranking. If a branch has fewer valid candidates than its quota, the vacant slots are filled by the highest-scoring unselected candidates from other branches. This fills the budget whenever the pooled candidate set contains at least $K$ valid items. Quota-based selection is retained before sorting; the procedure is not an unconstrained global top-$K$ search over all joint scores.

When reporting multiple cutoffs, we set the beam width and allocation budget to the largest cutoff $K_{\max}$ and construct one ranked list. Results at a smaller cutoff $K$ use the first $K$ entries of that list without reallocating quotas or repeating generation.

\section{Experiments}
\label{sec:experiments}
To evaluate \ourmodel, we conduct experiments to answer the following research questions (RQs):
\begin{itemize}[leftmargin=*]
    \item \textbf{RQ1:} How does \ourmodel perform against sequential and generative recommendation baselines?
    \item \textbf{RQ2:} How do linguistic encoding, Jev-like decision modeling, and length-aware inference contribute to performance?
    \item \textbf{RQ3:} Does separating the prediction heads alleviate gradient conflict between length and content prediction?
    \item \textbf{RQ4:} How efficient is \ourmodel in training and inference?
    \item \textbf{RQ5:} How sensitive is \ourmodel to the number of available lengths $N$ and the shortest SID length $M$?
    \item \textbf{RQ6:} How does frequency-aware encoding assign SID lengths to individual items in concrete examples?
\end{itemize}

\subsection{Experimental Setup}
\label{sec:experimental-setup}
\nosection{Datasets \& Benchmarks}
Following prior work~\citep{liu2025e2egrec,mtgrec}, we use three categories from Amazon Reviews 2023~\citep{hou2024bridging}: Musical Instruments (\textbf{Instrument}), Industrial \& Scientific (\textbf{Scientific}), and Video Games (\textbf{Game}). The review records cover May 1996 through September 2023. Following the preprocessing protocol in prior work~\citep{tiger,s3rec}, we exclude users and items with fewer than five interactions and sort the remaining interactions chronologically for each user. The maximum history length is 20 items rather than 20 tokens. For \ourmodel, each retained item is serialized as its length prefix followed by its SID, as defined in~\Cref{eq:history-serialization}. Item frequencies used for length assignment are computed only from the training split. Statistics of the adopted benchmarks are listed in~\Cref{tab:dataset-statistics}.
\begin{table}[t]
    \centering
    \caption{\textbf{Adopted Amazon Reviews benchmark statistics.} AvgLen is the average interaction sequence length.}
    \label{tab:dataset-statistics}
    \begingroup
    \small
    \setlength{\tabcolsep}{3pt}
    \begin{tabular}{lrrr}
        \toprule
        & \textbf{Instrument} & \textbf{Scientific} & \textbf{Game} \\
        \midrule
        Users & 57,439 & 50,985 & 94,762 \\
        Items & 24,587 & 25,848 & 25,612 \\
        Interactions & 511,836 & 412,947 & 814,586 \\
        Sparsity & 99.964\% & 99.969\% & 99.966\% \\
        AvgLen & 8.91 & 8.10 & 8.60 \\
        \bottomrule
    \end{tabular}
    \endgroup
\end{table}

\begin{table*}[t]
\setlength{\tabcolsep}{3pt}  %
\caption{\textbf{Comparison of \ourmodel with existing methods on three datasets.} Best results are \textbf{bold} and second-best results are \underline{underlined}. R@K and N@K denote Recall@K and NDCG@K, respectively. Improvements are relative to the best baseline in each column. ``$\ast$'' denotes statistically significant improvements ($p < 0.05$) over the best baseline according to a paired t-test.}
\label{table-RQ1}
\begin{center}
\resizebox{0.88\textwidth}{!}{%
\begin{tabular}{@{}lcccccccccccc@{}}
\toprule
\multicolumn{1}{c}{\multirow{2.5}{*}{\bfseries Methods}}
 & \multicolumn{4}{c}{Instrument} & \multicolumn{4}{c}{Scientific} & \multicolumn{4}{c}{Game} \\
\cmidrule(lr){2-5}\cmidrule(lr){6-9}\cmidrule(lr){10-13}
 & R@5 & R@10 & N@5 & N@10 & R@5 & R@10 & N@5 & N@10 & R@5 & R@10 & N@5 & N@10 \\
\midrule
Caser & 0.0241 & 0.0386 & 0.0151 & 0.0197 & 0.0159 & 0.0257 & 0.0101 & 0.0132 & 0.0330 & 0.0553 & 0.0209 &0.0281 \\
HGN & 0.0321 & 0.0517 & 0.0202 & 0.0265 & 0.0212 & 0.0351 & 0.0131 & 0.0176 & 0.0424 & 0.0687 & 0.0281 & 0.0356  \\
GRU4Rec & 0.0324 & 0.0501 & 0.0209 & 0.0266 & 0.0202 & 0.0338 & 0.0129 & 0.0173 & 0.0499 & 0.0799 & 0.0320 & 0.0416 \\
BERT4Rec & 0.0307 & 0.0485 & 0.0195 & 0.0252 & 0.0186 & 0.0296 & 0.0119 & 0.0155 & 0.0460 & 0.0735 & 0.0298 & 0.0386 \\
SASRec & 0.0333 & 0.0523 & 0.0213 & 0.0274 & 0.0259 & 0.0412 & 0.0150 & 0.0199 & 0.0535 & 0.0847 & 0.0331 & 0.0438 \\
FMLP-Rec & 0.0339 & 0.0536 & 0.0218 & 0.0282 & 0.0269 & 0.0422 & 0.0155 & 0.0204 & 0.0528 & 0.0857 & 0.0338 & 0.0444   \\
HSTU & 0.0343 & 0.0577 & 0.0191 & 0.0271 & 0.0271 & 0.0429 & 0.0147 & 0.0198 & 0.0578 & 0.0903 & 0.0334 &0.0442  \\
DuoRec &0.0347 &0.0547 &0.0227 &0.0291 &0.0234 &0.0389 &0.0146 &0.0196 & 0.0524&0.0827 &0.0336 & 0.0433 \\
FDSA & 0.0347 & 0.0545 & 0.0230 & 0.0293 & 0.0262 & 0.0421 & 0.0169 & 0.0213 &0.0544 &0.0852 & 0.0361 &0.0448\\
S\textsuperscript{3}-Rec
 & 0.0317 & 0.0496 & 0.0199 & 0.0257 & 0.0263 & 0.0418 & 0.0171 & 0.0219 &0.0485 &0.0769 &0.0315 &0.0406 \\
\midrule
TIGER & 0.0370 & 0.0564 & 0.0244 & 0.0306 & 0.0264 & 0.0422 & 0.0175 & 0.0226  & 0.0559 & 0.0868 & 0.0366 &0.0467 \\
LETTER & 0.0372 & 0.0580 & \underline{0.0246} & 0.0313 & 0.0279 & 0.0435 & \underline{0.0182} & 0.0232 & 0.0563& 0.0877& 0.0372& 0.0473 \\
ActionPiece  
&\underline{0.0383} &\underline{0.0615}  & 0.0243 &\underline{0.0318} 
&\underline{0.0284} &\underline{0.0452}  &\underline{0.0182} &\underline{0.0236} 
&\underline{0.0591} &\underline{0.0927}  &\underline{0.0382} &\underline{0.0490} 
\\
\midrule
\textbf{\ourmodel} & \textbf{0.0407}$^{\ast}$ & \textbf{0.0643}$^{\ast}$ & \textbf{0.0263}$^{\ast}$ & \textbf{0.0340}$^{\ast}$ & \textbf{0.0312}$^{\ast}$ & \textbf{0.0486}$^{\ast}$ & \textbf{0.0198}$^{\ast}$ & \textbf{0.0253}$^{\ast}$ & \textbf{0.0621}$^{\ast}$ & \textbf{0.0966}$^{\ast}$ & \textbf{0.0405}$^{\ast}$ & \textbf{0.0514}$^{\ast}$ \\
\midrule
Improvements & +6.27\% & +4.55\% & +6.91\% & +6.92\% & +9.86\% & +7.52\% & +8.79\% & +7.20\% & +5.08\% & +4.21\% & +6.02\% & +4.90\% \\
\bottomrule
\end{tabular}%
}
\end{center}
\end{table*}

\nosection{Evaluation Settings}
Following previous work~\citep{sasrec,tiger,s3rec}, we adopt leave-one-out splitting: each user's latest interaction is used for testing, the penultimate interaction for validation, and all earlier interactions for training. We evaluate against the full item catalog without negative sampling and report Recall@5, Recall@10, NDCG@5, and NDCG@10. Generative baselines use a beam width of 50. For \ourmodel, we follow~\Cref{sec:length-aware-inference} with a beam width of $K_{\max}=10$ in each length branch and construct one ranked list whose prefixes are used for both cutoffs.

\nosection{Compared Models}
We compare against 13 baselines in two categories. Among conventional sequential recommenders, Caser~\citep{caser} uses convolutional filters to model spatial and positional patterns in interaction sequences. HGN~\citep{hgn} models user preferences through feature-level and instance-level hierarchical gating. GRU4Rec~\citep{gru4rec} uses gated recurrent units to capture sequential dependencies in user behavior. BERT4Rec~\citep{bert4rec} learns bidirectional sequential representations through masked item prediction. SASRec~\citep{sasrec} uses unidirectional self-attention to model dependencies in interaction histories. FMLP-Rec~\citep{fmlprec} combines an MLP-based architecture with learnable filters for sequence denoising. HSTU~\citep{hstu} uses hierarchical sequential transducers to model user actions and their timestamps at scale. DuoRec~\citep{duorec} addresses representation degeneration through contrastive regularization with dropout-based augmentation and supervised positive sampling. FDSA~\citep{fdsa} uses separate self-attention streams to model feature-level and item-level dependencies. S3-Rec~\citep{s3rec} learns from correlations among items and their attributes through self-supervised objectives.

For generative recommendation, TIGER~\citep{tiger} uses RQ-VAE to convert item representations into semantic IDs and predicts the next item's ID autoregressively. LETTER~\citep{letter} extends learnable item tokenization with collaborative information and diversity-oriented constraints. ActionPiece~\citep{actionpiece} tokenizes action sequences by merging co-occurring features in context and uses set permutation regularization to model feature sets.

\nosection{Implementation Details}
We use a T5-style encoder--decoder with four layers in each stack and a hidden size of 128. We use two H20 GPUs with a per-GPU batch size of 256 and train for 200 epochs on each dataset. Optimization uses AdamW with a cosine learning-rate schedule. The learning rate is selected from $\{0.01,0.003\}$ and weight decay from $\{0.1,0.05,0.035\}$. The length and content objectives have unit weights and are optimized in one joint stage, following~\Cref{sec:joint-training}. Further details of the item tokenizer, architecture, and decoding are provided in~\Cref{app:implementation}.

\subsection{Overall Performance (RQ1)}
\label{sec:rq1}

We compare \ourmodel with conventional sequential and generative recommendation methods on the three datasets. The results in~\Cref{table-RQ1} lead to the following observations.

\begin{itemize}[leftmargin=*]
    \item \textbf{Obs 1: \ourmodel consistently outperforms both families of baselines.} Our method achieves the best result in all twelve dataset--metric combinations, with a mean relative improvement of 6.52\% over the strongest baseline in each column. The advantage holds across all three item domains and both Recall and NDCG, showing that the overall quality gain is not confined to a particular dataset or evaluation criterion.

    \item \textbf{Obs 2: The gains extend beyond the standard semantic-ID pipeline.} Compared with TIGER's fixed-length SIDs, \ourmodel improves every metric with a mean relative gain of 12.04\% across the reported settings. The advantage also holds over LETTER and ActionPiece, showing that it persists with stronger tokenization baselines. These results support frequency-aware SID encoding with explicit length decisions.

    \item \textbf{Obs 3: The gains persist at the top of the recommendation list.} The relative Recall@5 improvement over the best baseline exceeds its Recall@10 counterpart on every dataset. NDCG also improves at both cutoffs, so the gains are not limited to retrieving relevant items when more output slots are available. This is particularly relevant to short recommendation lists, where the quality of the highest-ranked predictions matters most.
\end{itemize}

\subsection{Ablation Study (RQ2)}
\label{sec:rq2}
\begin{table*}[t]
    \centering
    \caption{\textbf{Ablation study of \ourmodel on three datasets.} R@K and N@K denote Recall@K and NDCG@K.}
    \label{table-RQ2}
    \begingroup
    \setlength{\tabcolsep}{3pt}
    \resizebox{0.88\textwidth}{!}{%
    \begin{tabular}{@{}lcccccccccccc@{}}
        \toprule
        \multirow{2}{*}{\textbf{Variant}} & \multicolumn{4}{c}{Instrument} & \multicolumn{4}{c}{Scientific} & \multicolumn{4}{c}{Game} \\
        \cmidrule{2-13}
        & R@5 & R@10 & N@5 & N@10 & R@5 & R@10 & N@5 & N@10 & R@5 & R@10 & N@5 & N@10 \\
        \midrule
        \multicolumn{1}{c}{} & \multicolumn{12}{c}{\textit{Linguistic Encoding}} \\
        \cmidrule[\lightrulewidth]{2-13}
        (1.1) Fixed-length & 0.0370 & 0.0564 & 0.0244 & 0.0306 & 0.0264 & 0.0422 & 0.0175 & 0.0226 & 0.0559 & 0.0868 & 0.0366 & 0.0467 \\
        (1.2) Random lengths & 0.0374 & 0.0591 & 0.0239 & 0.0311 & 0.0278 & 0.0445 & 0.0178 & 0.0230 & 0.0567 & 0.0898 & 0.0369 & 0.0476 \\
        (1.3) Reversed frequency & 0.0358 & 0.0579 & 0.0229 & 0.0300 & 0.0270 & 0.0437 & 0.0171 & 0.0225 & 0.0552 & 0.0885 & 0.0356 & 0.0462 \\
        \midrule
        \multicolumn{1}{c}{} & \multicolumn{12}{c}{\textit{Jev-like Decision Modeling}} \\
        \cmidrule[\lightrulewidth]{2-13}
        (2.1) Shared LM head & 0.0377 & 0.0598 & 0.0245 & 0.0315 & 0.0282 & 0.0443 & 0.0184 & 0.0234 & 0.0574 & 0.0910 & 0.0370 & 0.0481 \\
        (2.2) w/o Length tokens & 0.0320 & 0.0501 & 0.0205 & 0.0260 & 0.0224 & 0.0358 & 0.0132 & 0.0178 & 0.0508 & 0.0809 & 0.0321 & 0.0419 \\
        (2.3) History-only length head & 0.0405 & 0.0640 & 0.0261 & 0.0338 & 0.0310 & 0.0489 & 0.0197 & 0.0255 & 0.0623 & 0.0963 & 0.0407 & 0.0512 \\
        \midrule
        \multicolumn{1}{c}{} & \multicolumn{12}{c}{\textit{Length-aware Inference}} \\
        \cmidrule[\lightrulewidth]{2-13}
        (3.1) w/o Length quotas & 0.0397 & 0.0631 & 0.0255 & 0.0331 & 0.0302 & 0.0475 & 0.0191 & 0.0246 & 0.0612 & 0.0953 & 0.0397 & 0.0506 \\
        (3.2) w/o Length-aware ranking & 0.0385 & 0.0643 & 0.0244 & 0.0323 & 0.0288 & 0.0486 & 0.0180 & 0.0236 & 0.0595 & 0.0966 & 0.0382 & 0.0494 \\
        \midrule
        \ourmodel & 0.0407 & 0.0643 & 0.0263 & 0.0340 & 0.0312 & 0.0486 & 0.0198 & 0.0253 & 0.0621 & 0.0966 & 0.0405 & 0.0514 \\
        \bottomrule
    \end{tabular}%
    }
    \endgroup
\end{table*}

We ablate the three stages of \ourmodel and retrain each variant, except that the two inference variants reuse the trained model of \ourmodel and its candidate pool. For linguistic encoding, Fixed-length reuses the TIGER results, Random lengths draws each item's SID length at random from the admissible range independently of its frequency, and Reversed frequency assigns long SIDs to popular items and short SIDs to long-tail items. For Jev-like decision modeling, Shared LM head predicts the length token and the SID with a single autoregressive output head, w/o Length tokens removes the length tokens from both the history and the target, and History-only length head replaces the candidate-conditioned scorer with a classifier on the start-position state. For length-aware inference, w/o Length quotas ranks all valid candidates by the joint score without per-length quotas, and w/o Length-aware ranking keeps the items selected by \ourmodel but orders them by the conditional generation probability alone. The results in~\Cref{table-RQ2} lead to the following observations.

\begin{itemize}[leftmargin=*]
    \item \textbf{Obs 4: The encoding gain comes from tying SID length to interaction frequency.} All three encoding variants fall below \ourmodel on every metric. Random lengths retains variable-length SIDs and the complete decision pipeline but still trails \ourmodel by 8.70\% on average, so most of the encoding gain stems from assigning lengths by frequency rather than from variable length itself. Reversing the relation is even more harmful: Reversed frequency is worse than Random lengths on every metric and falls below Fixed-length on all NDCG metrics, which supports giving popular items short codes and long-tail items longer descriptions as natural language does.

    \item \textbf{Obs 5: The benefit of Jev-like decision modeling comes from an explicit length decision rather than a particular classifier.} Shared LM head lowers every metric by 7.50\% on average, showing that predicting the length through the same output head as the content tokens weakens recommendation even when the encoding is unchanged. Removing the length tokens from both the history and the target causes the largest average degradation in this group, since the model must then infer item boundaries and the target length from the SID stream alone. In contrast, replacing the candidate-conditioned scorer with a history-only classifier keeps every metric within 1\% of \ourmodel, so the gain stems from separating the length decision from content generation and linking the two through length tokens rather than from a specific scorer design.

    \item \textbf{Obs 6: Length probabilities contribute to both candidate allocation and cross-length ranking.} Although both inference variants use the same trained model as \ourmodel, removing length quotas lowers all twelve metrics. Removing length-aware ranking leaves Recall@10 unchanged because the selected items are identical by construction, but it lowers Recall@5 and every NDCG metric, with larger NDCG drops than removing quotas on every dataset. Quotas therefore keep the list composition consistent with the predicted length distribution, while length-aware ranking ranks items of more probable lengths higher.
\end{itemize}

\subsection{Optimization Conflict Analysis (RQ3)}
\label{sec:rq3}

\begin{figure}[!htbp]
    \centering
    \includegraphics[width=\columnwidth]{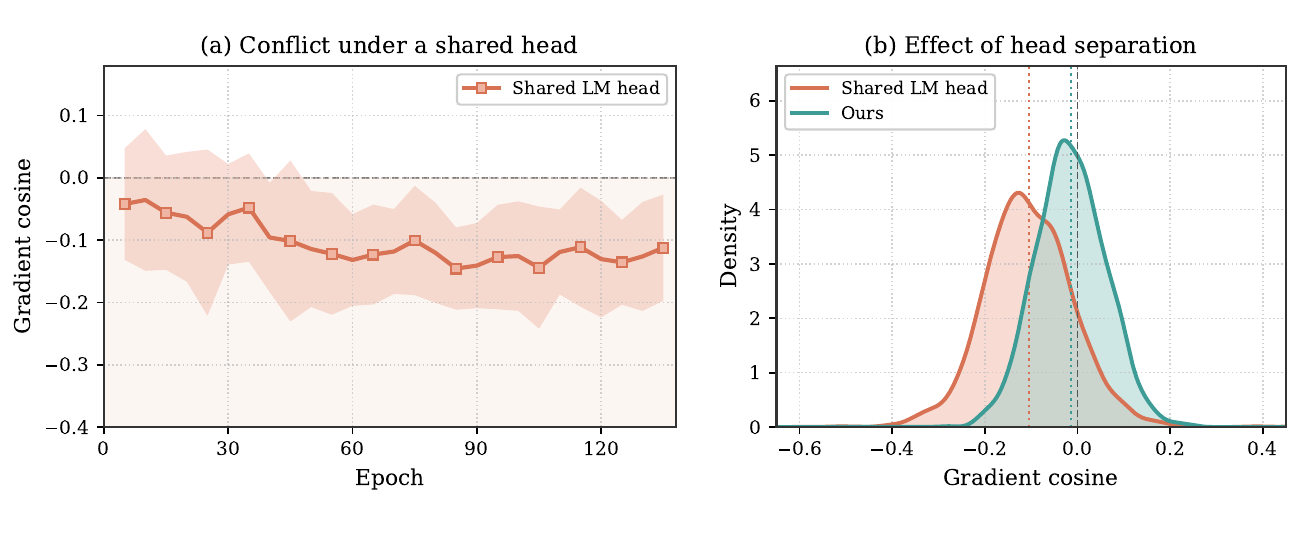}
    \caption{\textbf{Length and content gradient alignment.} (a) Shared-head batch mean with standard deviation. (b) Pooled cosine densities; dotted lines mark means.}
    \label{fig:rq3-conflict}
    \Description{The left panel shows the batch-mean gradient cosine for the shared LM head across epochs 5 to 135, with one standard deviation shading. The mean stays negative at every recorded checkpoint. The right panel compares pooled cosine densities for the shared head and our method; our distribution moves closer to zero but retains negative samples.}
\end{figure}

We compare the Shared LM head variant with \ourmodel using the batch-wise cosine similarity between the gradients of the length and content losses on a common set of shared decoder parameters. Negative cosine indicates conflicting directions. We use 48 batch measurements at each of 27 checkpoints from epoch 5 to 135. We track the shared-head mean over epochs and pool the measurements from this window for the distribution comparison in~\Cref{fig:rq3-conflict}.

\begin{itemize}[leftmargin=*]
    \item \textbf{Obs 7: Length and content objectives exhibit persistent gradient conflict under a shared head.} The mean gradient cosine is negative at every recorded checkpoint. It becomes more negative over the earlier part of training and then fluctuates at a negative level rather than returning to alignment. This shows that the two objectives continue to compete in their updates to the measured shared parameters instead of resolving their directional mismatch through training alone.

    \item \textbf{Obs 8: Separating the prediction heads alleviates gradient conflict.} The cosine distribution of \ourmodel shifts toward zero, with its pooled mean increasing from the shared-head value of $-0.105$ to $-0.015$. Negative samples become less prevalent, indicating weaker directional interference rather than complete independence. These results support task-specific output modeling as a way to reduce conflict between length decisions and content generation while retaining a shared backbone.
\end{itemize}

\subsection{Cost Efficiency Analysis (RQ4)}
\label{sec:rq4}

\begin{figure}[!htbp]
    \centering
    \includegraphics[width=\columnwidth]{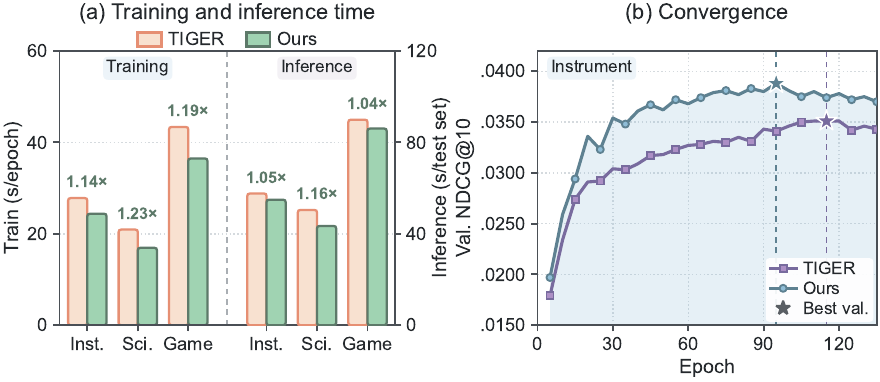}
    \caption{\textbf{Runtime and validation progress.} (a) Training and inference time with baseline-to-ours ratios. (b) Instrument validation NDCG@10; stars mark the best checkpoints.}
    \label{fig:rq4-efficiency}
    \Description{Panel a compares mean training seconds per epoch and total inference seconds on each full test set for Instrument, Scientific, and Game. Our method uses less time in all six comparisons. Panel b shows validation NDCG at 10 on Instrument, with the best recorded checkpoints at epoch 95 for our method and epoch 115 for TIGER.}
\end{figure}

We compare the efficiency of \ourmodel and TIGER in~\Cref{fig:rq4-efficiency}. We measure average training time per epoch and total inference time over the complete test set of each dataset, and track validation NDCG@10 across training epochs on Instrument.

\begin{itemize}[leftmargin=*]
    \item \textbf{Obs 9: \ourmodel reduces both training and inference time.} Training is faster on all three datasets, with a mean relative reduction of 15.75\% in time per epoch. Total test-set inference time also decreases on every dataset. These savings are consistent with assigning compact codes to frequently occurring items and reusing the history representation for length decisions and content generation. Together with the recommendation gains in~\Cref{table-RQ1}, the results show that \ourmodel improves accuracy while lowering runtime under the evaluated settings.

    \item \textbf{Obs 10: \ourmodel reaches a higher validation peak in fewer epochs.} On Instrument, \ourmodel achieves higher validation NDCG@10 at every recorded checkpoint and reaches its best score 20 epochs earlier than TIGER. The advantage therefore extends beyond faster execution within an epoch to stronger learning progress across epochs. This complements the per-epoch savings: \ourmodel requires less computation time per epoch and fewer epochs to reach its observed validation optimum.
\end{itemize}

\subsection{Hyperparameter Analysis (RQ5)}
\label{sec:rq5}

\begin{figure}[!htbp]
    \centering
    \includegraphics[width=\columnwidth]{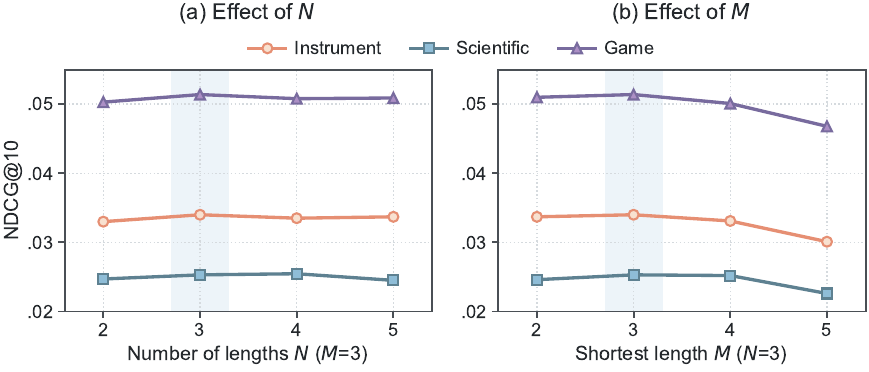}
    \caption{\textbf{SID length sensitivity.} (a) Vary $N$ at $M=3$. (b) Vary $M$ at $N=3$. Shading marks the default value.}
    \label{fig:rq5-hyperparameters}
    \Description{NDCG at 10 on Instrument, Scientific, and Game for separate sweeps of N and M from 2 to 5. The left panel varies N at fixed M=3; the right panel varies M at fixed N=3. The shaded bands indicate the default value 3, not necessarily the best score.}
\end{figure}

We study the sensitivity of \ourmodel to the number of available SID lengths $N$ and the shortest SID length $M$. We vary each parameter over $\{2,3,4,5\}$ while fixing the other at its default value of 3. The NDCG@10 results on all three datasets are shown in~\Cref{fig:rq5-hyperparameters}.

\begin{itemize}[leftmargin=*]
    \item \textbf{Obs 11: \ourmodel is robust to moderate changes in the length configuration.} Performance remains close to the default across the full $N$ sweep and for $M\in\{2,3,4\}$. In these settings, NDCG@10 differs from the default by less than 3.2\% on every dataset. This consistent behavior suggests that \ourmodel does not depend on precise tuning near the default configuration.

    \item \textbf{Obs 12: Moderate length settings work best, while increasing the minimum length too far hurts performance.} A value of 3 gives the best result in most dataset--parameter combinations. Increasing $N$ beyond 3 produces small, non-monotonic changes rather than consistent improvements. In contrast, increasing $M$ beyond 3 lowers NDCG@10 on all three datasets, with a sharper decline from 4 to 5. Raising $M$ lengthens the codes of every frequency group, including popular items that would otherwise receive compact identifiers. The results therefore favor moderate minimum lengths rather than uniformly adding semantic tokens to all items.
\end{itemize}

\subsection{Case Study (RQ6)}
\label{sec:rq6}

\begin{figure}[!htbp]
    \centering
    \includegraphics[width=\columnwidth]{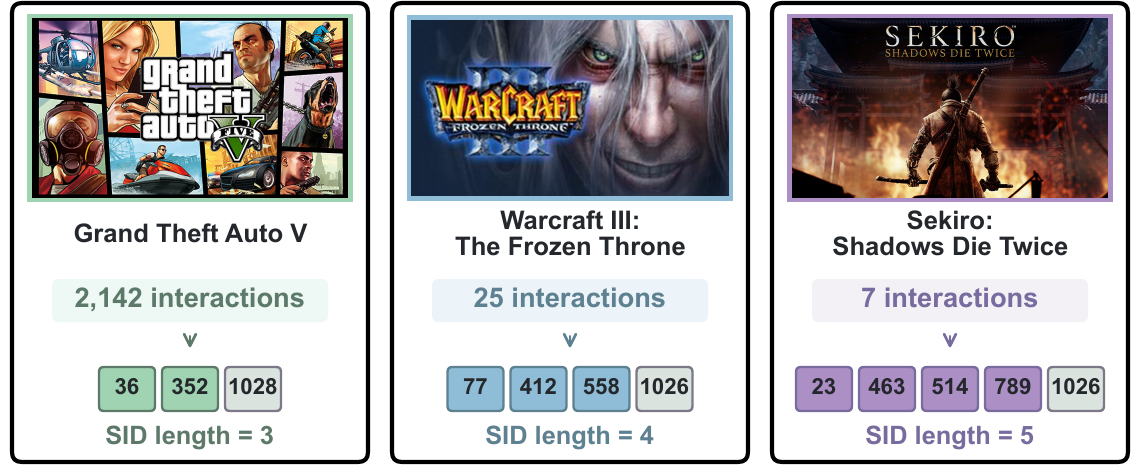}
    \caption{\textbf{Frequency-aware SID lengths for three game items.} Each card shows the item's interaction count and complete SID, including its final collision token.}
    \label{fig:rq6-case-study}
    \Description{Three game items with different interaction counts and SID lengths. Grand Theft Auto V has 2,142 interactions and SID (36, 352, 1028), of length 3. Warcraft III: The Frozen Throne has 25 interactions and SID (77, 412, 558, 1026), of length 4. Sekiro: Shadows Die Twice has 7 interactions and SID (23, 463, 514, 789, 1026), of length 5.}
\end{figure}

To illustrate how \ourmodel adapts item representations to interaction frequency, we examine tokenization for three Game items at the default $M=N=3$, as shown in~\Cref{fig:rq6-case-study}.

The three games differ substantially in interaction frequency and consequently receive SIDs of different lengths. Specifically, \emph{Grand Theft Auto V} has 2,142 interactions and is encoded as \texttt{[36,\allowbreak352,\allowbreak1028]}. \emph{Warcraft III: The Frozen Throne} has 25 interactions and receives the longer SID \texttt{[77,\allowbreak412,\allowbreak558,\allowbreak1026]}. With only seven interactions, \emph{Sekiro: Shadows Die Twice} receives the five-token SID \texttt{[23,\allowbreak463,\allowbreak514,\allowbreak789,\allowbreak1026]}. Since each SID includes one final collision-resolution token, the additional positions retain more semantic codes from the shared residual quantizer. This case study illustrates the linguistic encoding principle: frequently encountered items use compact identifiers, whereas less frequent items retain richer semantic descriptions to complement their sparse interaction signals.

\section{Conclusion}
\label{sec:conclusion}

We presented \ourmodel, a generative recommendation framework inspired by compact frequent expressions and contextual generation in natural language. Linguistic encoding assigns shorter SIDs to popular items and richer semantic descriptions to long-tail items while preserving shared semantics and unique item identities through shared codebooks and collision-resolution tokens. This frequency-aware allocation reduces the encoding cost repeatedly incurred by frequent interactions without requiring all items to use equally short representations. Jev-like decision modeling uses separate length and content heads that share a representation of user history and interact through length-conditioned generation. Length-aware inference uses the predicted length probabilities to guide candidate allocation and cross-length ranking, integrating decisions across decoding branches rather than committing to a single predicted length. Experiments on three public datasets show consistent gains over the compared baselines and lower training and inference time than TIGER. Ablations support the component contributions, and gradient analysis shows that separate heads alleviate conflict between the two objectives. These findings support treating SID length as both a representation choice and an explicit prediction decision for accurate and efficient generative recommendation.

\bibliographystyle{ACM-Reference-Format}
\bibliography{reference}

\appendix
\crefalias{section}{appendix}
\crefalias{subsection}{appendix}
\crefalias{subsubsection}{appendix}
\section{Experimental Details}
\label{app:experimental-details}

We follow the same experimental settings as prior work~\citep{zhong2025pctx} and directly adopt its reported results for most baselines.

\subsection{Implementation Details.}
\label{app:implementation}
\begin{itemize}
    \item \textbf{Item Tokenizer.} Following TIGER~\citep{tiger}, we encode item textual attributes with \texttt{sentence-t5-base}~\citep{sentence-t5}. We apply PCA with whitening~\citep{su2021whitening} before fitting a residual quantizer with FAISS~\citep{faiss}. Each semantic codebook contains 256 entries. The quantizer uses $M+N-2$ levels, and each item retains a frequency-dependent prefix followed by a collision-resolution token, as specified in~\Cref{sec:linguistic-encoding}. The semantic codebook size does not prescribe the capacity of the separate collision-token vocabulary.

    \item \textbf{Recommendation Backbone.} The encoder and decoder each contain four layers. The hidden size is 128, the feed-forward inner dimension is 512, and each attention module has four heads with a per-head dimension of 64. We use ReLU activations. The length head and content head share this backbone according to~\Cref{sec:decision-modeling}.

    \item \textbf{Prediction Heads.} We retain the native T5 content head: a single bias-free linear projection from 128 dimensions to the token vocabulary, with weights tied to the token embeddings. The added length head concatenates the 128-dimensional START state and the 128-dimensional candidate length embedding, then scores each candidate with a shared $256\to256\to128\to1$ MLP using GELU after the first two linear layers.

    \item \textbf{Decoding Configuration.} We use a beam width of 50 for generative baselines. For \ourmodel, the evaluation cutoffs give $K_{\max}=10$, which is both the beam width in each length branch and the total recommendation-list budget. We allocate slots and merge candidates once at this cutoff; Recall@5 and NDCG@5 are computed from the first five entries of the same list. The probabilities include EOS and are not length-normalized.

    \item \textbf{Method-Specific Hyperparameters.} On all datasets we set $M=N=3$, giving SID lengths of $3$, $4$, and $5$ with a four-level shared residual quantizer.
\end{itemize}

\end{document}